\documentclass{article} 
\usepackage{iclr2027_conference,times}
\iclrfinalcopy

\usepackage{amsmath,amsfonts,bm}

\def\eqref#1{equation~\ref{#1}}

\def\1{\bm{1}}

\DeclareMathAlphabet{\mathsfit}{\encodingdefault}{\sfdefault}{m}{sl}
\SetMathAlphabet{\mathsfit}{bold}{\encodingdefault}{\sfdefault}{bx}{n}

\usepackage{graphicx}
\usepackage{tikz}
\usetikzlibrary{positioning}
\usepackage{forest}
\usepackage{multirow}
\usepackage{booktabs}
\usepackage{caption}
\usepackage{hyperref}
\hypersetup{
    colorlinks=true,
    citecolor=blue,  
    linkcolor=blue,  
    urlcolor=blue    
}
\usepackage{algorithm}
\usepackage{algorithmic}
\usepackage{url}
\definecolor{myblue}{RGB}{80,150,255}
\definecolor{mygreen}{RGB}{100,200,120}
\definecolor{myyellow}{RGB}{255,210,80}
\definecolor{myred}{RGB}{240,100,100}

\title{Principles that Guide, Actions that Inform:\\
Agent Evolution via Knowledge Abstraction}

\author{
Bowen Ye$^{1,2}$,
Yongchao Xu$^{1,3}$,
Junkai Ma$^{1}$\thanks{Corresponding authors.},
Xiang Yin$^{2}$\footnotemark[1],
Wenzhao Li$^{1}$\thanks{Project leader.}
\\
$^{1}$Alibaba Group, Hangzhou, China\\
$^{2}$School of Automation and Intelligent Sensing, Shanghai Jiao Tong University, Shanghai 200240, China\\
$^{3}$University of Science and Technology of China, Hefei, China
}

\begin{document}

%

\maketitle

\begin{abstract}
Large language model (LLM) agents have demonstrated strong capabilities in interactive environments, yet their ability to continually evolve from experience remains limited. Although fine-tuning provides a pathway for adaptation, its reliance on parameter access and substantial computational resources limits its flexibility and scope of application, particularly for large-scale and closed-source LLMs. External memory provides an alternative by enabling agents to accumulate experience without modifying model parameters. However, existing approaches primarily focus on how experiences are represented and organized, while the acquired knowledge often remains closely coupled with specific tasks and contexts, limiting its generalization for continual agent evolution. A fundamental challenge is therefore how to transform concrete interactions into abstract and reusable knowledge that can guide future decisions and enable agents to generalize beyond individual experiences.

To address this issue, we propose SAGA
(\underline{\textbf{S}}elf-evolving
\underline{\textbf{A}}gents through
Experience-\underline{\textbf{G}}rounded
\underline{\textbf{A}}bstraction),
a framework for experience-grounded knowledge abstraction and
utilization in LLM agents.
SAGA progressively transforms interaction trajectories into
episodic descriptions, reusable procedures, and principles with
explicit applicability conditions, while preserving links to
their supporting execution evidence.
To make abstract knowledge actionable, SAGA instantiates
retrieved principles into task-specific operational guidance
and uses principle-derived checks to refine candidate actions
through corrective feedback and resampling.
These mechanisms form an execution--abstraction feedback loop,
in which accumulated knowledge guides subsequent interactions
and newly collected experiences update the hierarchical memory.
Experiments on ScienceWorld and ALFWorld demonstrate improved
task performance, while ablation studies show that contextual
instantiation and action regulation are important for realizing
the benefits of principle-level knowledge.

\end{abstract}

\section{Introduction}

The rapid advancement of large language models (LLMs) has enabled a new paradigm of autonomous agents, where pretrained models are augmented with external tools, memory modules, and interaction interfaces for complex reasoning and decision-making tasks. 
Recent LLM-based agents have demonstrated promising capabilities in diverse interactive environments, including embodied systems~\citep{brohan2023can,wang2023voyager}, text-based environments such as ALFWorld and ScienceWorld~\citep{shridhar2020alfworld,wang2022scienceworld}, and emerging computer-use scenarios. 
However, existing agents largely rely on the capabilities acquired during pretraining and lack mechanisms for continual adaptation after deployment. 
A fundamental challenge toward more capable agents is therefore enabling self-evolution, where agents can learn from interaction experiences, refine their behaviors, and gradually acquire task-specific or user-specific knowledge beyond the original foundation model.

Recent studies have explored several directions toward self-evolving agents. 
One direction adapts the model itself through supervised fine-tuning or reinforcement learning, enabling direct modification of agent behaviors but requiring substantial computational resources and access to model parameters~\citep{shao2024deepseekmath,schulman2017proximal}. 
Another direction performs adaptation at inference time through reasoning, reflection, and self-improvement mechanisms without updating model parameters~\citep{yao2022react,madaan2023self,shinn2023reflexion}. 
Recent works further investigate experience-driven adaptation by extracting reusable skills or strategies from previous interactions~\citep{wang2023voyager,zhao2024expel,xue2026strata}. 
Among these approaches, external memory provides an effective mechanism for continual adaptation of LLM agents. 
Existing memory systems study persistent memory management~\citep{packer2023memgpt,zhong2024memorybank}, structured memory representations~\citep{gutierrez2024hipporag,rasmussen2025zep,xu2026mem}, and adaptive memory evolution~\citep{zhang2026agentic,fan2026evolving,cao2026remember,zhang2026memrl}. 
These methods demonstrate the importance of memory as an interface between foundation models and evolving environments.

Despite recent progress in agent memory systems, existing approaches primarily focus on storing and retrieving information from past interactions. Memory architectures such as MemGPT and MemoryBank maintain persistent interaction histories for long-term context reuse \citep{packer2023memgpt,zhong2024memorybank}, while retrieval-based methods further improve access to stored experiences through structured memory mechanisms \citep{gutierrez2024hipporag}. However, these memories are typically represented as episodic records, retrieved contexts, or task-specific procedures, providing limited support for extracting transferable knowledge across tasks and environments. Recent experience-driven approaches, including Reflexion, ExpeL, and Voyager, begin to explore how accumulated experiences can improve future agent behavior through reflection, skill acquisition, and experience reuse \citep{shinn2023reflexion,zhao2024expel,wang2023voyager}. Nevertheless, how to systematically transform interaction experiences into higher-level, experience-grounded knowledge that captures reusable principles and applicability conditions remains a fundamental challenge for lifelong adaptation of LLM-based agents \citep{zheng2025towards,zhang2026experience}.

\begin{figure}
    \centering
    \includegraphics[width=0.925\linewidth]{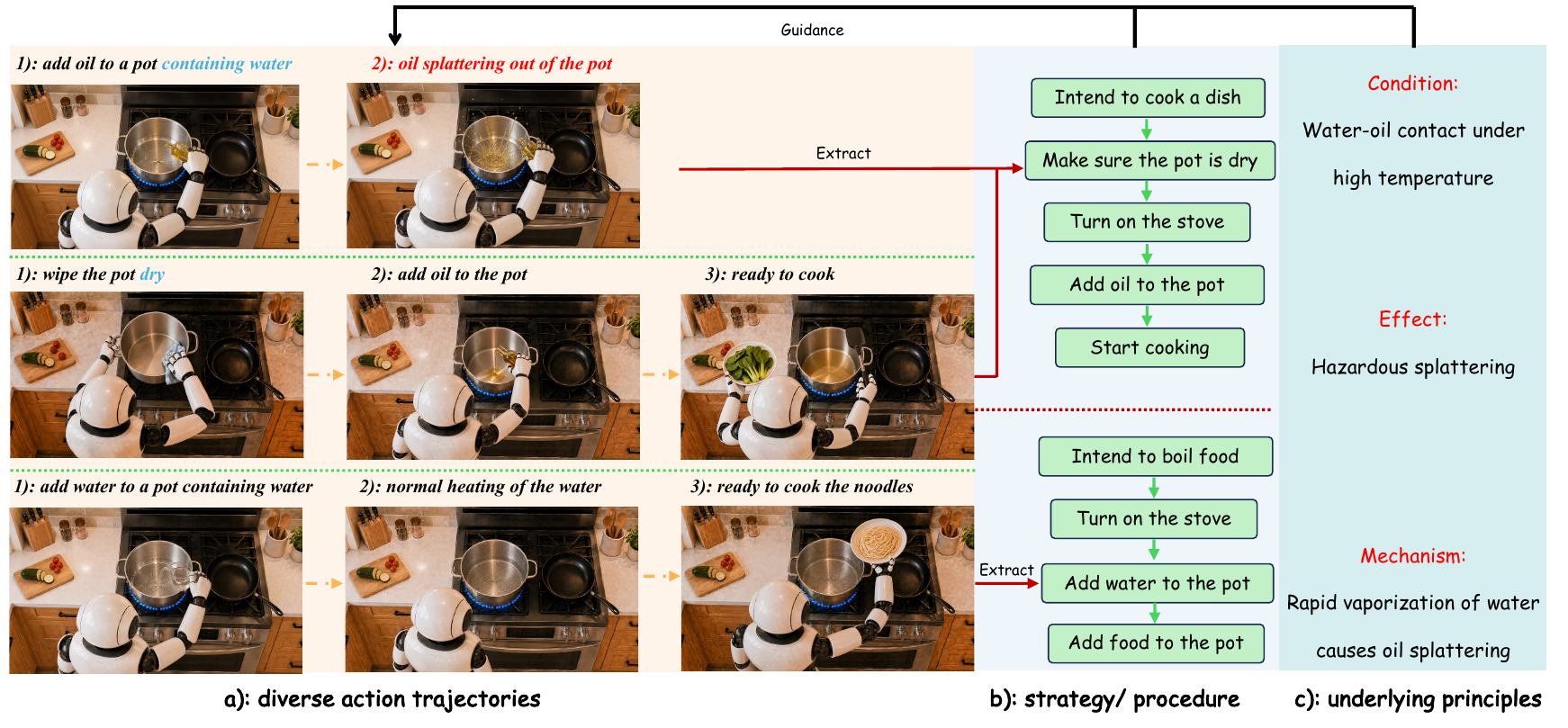}
    \vspace{-1.1em}
    \caption{Illustrative Example of Experience-Grounded Knowledge Evolution. 
    Interaction experiences are abstracted into reusable knowledge, which guides future actions.}
    \vspace{-1.5em}
    \label{fig:motivation}
\end{figure}

To address this challenge, we propose SAGA
(\underline{\textbf{S}}elf-evolving
\underline{\textbf{A}}gents through
Experience-\underline{\textbf{G}}rounded
\underline{\textbf{A}}bstraction),
a framework for experience-driven agent adaptation through
evidence-linked knowledge abstraction and utilization.
As illustrated in Fig.~\ref{fig:motivation}, SAGA establishes
an execution--abstraction feedback loop: accumulated knowledge
guides subsequent interactions, while newly collected trajectories
provide evidence for updating the knowledge representation.
The framework organizes experiences into four abstraction levels,
from raw trajectories and episodic descriptions to reusable
procedures and principles with explicit applicability conditions.
Links between adjacent levels preserve the connection between
abstract knowledge and its supporting execution evidence.
For knowledge utilization, SAGA instantiates retrieved principles
into operational guidance conditioned on the current task and
environmental context.
Principle-derived checks further regulate candidate actions
through corrective feedback and bounded resampling.
These mechanisms connect knowledge abstraction with both
task-level guidance and execution-time action refinement.
We evaluate SAGA on ScienceWorld~\citep{wang2022scienceworld}
and ALFWorld~\citep{shridhar2020alfworld} under memory-based
and parameter-based adaptation settings.
The results demonstrate improved task performance, while
ablation studies identify contextual instantiation and action
regulation as complementary mechanisms for effective
principle-level knowledge reuse.

Our contributions are summarized as follows.
\textbf{i)} We develop an evidence-linked knowledge abstraction
pipeline that organizes interaction experiences into raw
trajectories, episodic knowledge, reusable procedures, and
principles with explicit applicability conditions, preserving
traceability across abstraction levels.
\textbf{ii)} We propose SAGA, which integrates task-conditioned
principle instantiation and principle-based action regulation
within an execution--abstraction feedback loop, connecting
experience-derived knowledge with operational guidance and
action refinement.
\textbf{iii)} We evaluate SAGA on ScienceWorld and ALFWorld
under both memory-based and parameter-based adaptation settings,
and analyze the contributions of knowledge abstraction,
contextual instantiation, and action regulation through
dual-backbone ablation studies.

\section{Related Work}

\textbf{Self-Evolving LLM Agents.}
LLM agents increasingly incorporate feedback and accumulated experience
to improve reasoning and decision making.
ReAct~\citep{yao2022react} interleaves reasoning with environmental actions,
while Self-Refine~\citep{madaan2023self} and
Reflexion~\citep{shinn2023reflexion} use iterative feedback and reflection
to improve generated responses and subsequent task execution.
Beyond individual attempts, Voyager~\citep{wang2023voyager}
accumulates executable skills for open-ended exploration, and
ExpeL~\citep{zhao2024expel} combines retrieved experiences with insights
extracted from successful and unsuccessful trajectories.
StraTA~\citep{xue2026strata} jointly optimizes strategy generation
and action execution through agentic reinforcement learning.
Evolving-RL~\citep{fan2026evolving}, MemRL~\citep{zhang2026memrl},
and Remember Me, Refine Me~\citep{cao2026remember} further explore
complementary mechanisms for experience-driven improvement.
These studies motivate agent adaptation through both external knowledge
and parameter updates.
SAGA focuses on connecting experience abstraction with knowledge
utilization within a recurring execution--abstraction loop.

\textbf{Memory for Continual Agent Adaptation.}
External memory allows agents to retain information across interactions
without requiring updates to the underlying model parameters.
MemGPT~\citep{packer2023memgpt} introduces memory management inspired
by operating systems, while MemoryBank~\citep{zhong2024memorybank}
supports long-term memory for personalized interactions.
Subsequent studies investigate graph-based
retrieval~\citep{gutierrez2024hipporag},
temporal memory architectures~\citep{rasmussen2025zep},
and agentic memory construction~\citep{xu2026mem}.
Mem0~\citep{chhikara2025mem0} explores scalable memory management,
while agentic context engineering~\citep{zhang2026agentic}
treats context as an evolving resource.
CoEvo-Mem~\citep{ye2026coevo} further studies the joint evolution
of retrieval policies and memory banks.
These approaches provide complementary mechanisms for maintaining,
organizing, and updating persistent knowledge.
SAGA places particular emphasis on the relationship between abstraction
levels and their execution evidence, and on converting retrieved
principles into operational guidance and action checks.

\textbf{Experience Abstraction and Knowledge Utilization.}
Several studies directly investigate how interaction experience
can be transformed into reusable knowledge.
Voyager~\citep{wang2023voyager} represents acquired behaviors
as executable skills, and ExpeL~\citep{zhao2024expel}
extracts cross-task insights from prior trajectories.
CLIN~\citep{majumder2023clin} maintains a dynamic textual memory
of causal abstractions and updates it across trials to support
task adaptation and generalization.
AutoManual~\citep{chen2024automanual} constructs instruction manuals
through online rule generation and revision, retaining examples
and validation records associated with the rules.
AutoGuide~\citep{fu2024autoguide} extracts conditional guidelines
from offline experiences and selects relevant guidance according
to the current interaction context.
The experience compression perspective~\citep{zhang2026experience}
further relates memories, skills, and rules through different
degrees of abstraction.
Building on these advances, SAGA integrates knowledge abstraction
and execution-time utilization through three connected mechanisms:
an evidence-linked hierarchy spanning trajectories, episodes,
procedures, and principles; contextual instantiation that converts
retrieved principles into task-specific operational guidance;
and principle-derived executable checks that refine candidate
actions through corrective feedback and bounded resampling.
Where AutoManual~\citep{chen2024automanual} emphasizes rule management
and manual construction, and AutoGuide~\citep{fu2024autoguide}
emphasizes context-aware guideline selection, SAGA explicitly
couples a multi-level abstraction pipeline with post-retrieval
instantiation and action regulation.
New execution trajectories subsequently provide evidence for
updating the hierarchy, closing the loop between knowledge
construction and utilization.

\section{SAGA: Experience-Grounded Knowledge Evolution}
\begin{figure}
    \centering
    \includegraphics[width=0.99\linewidth]{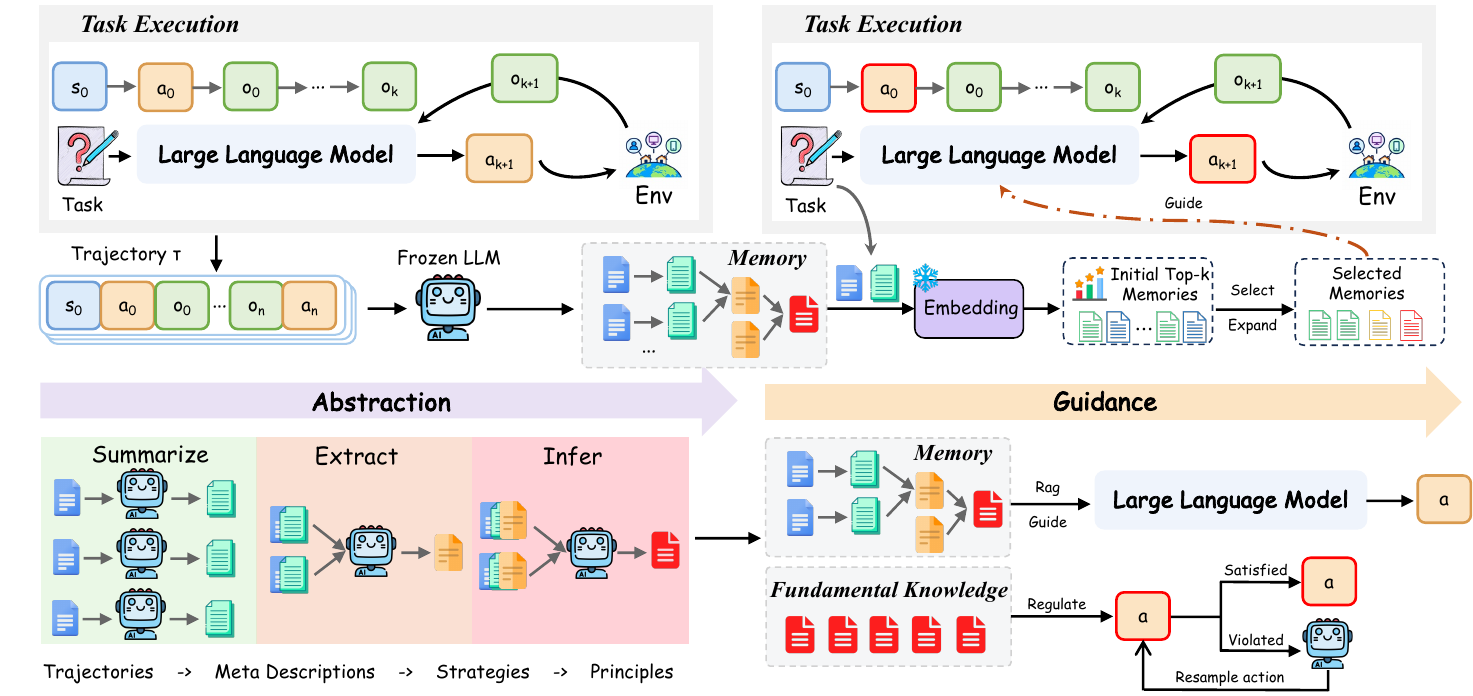}
    \caption{Overview of SAGA, illustrating the close-loop evolving process from experience abstraction to knowledge-guided decision making.}
    \vspace{-1.5em}
    \label{fig:framework}
\end{figure}
\subsection{Overview}
In this work, we propose SAGA (\underline{\textbf{S}}elf-evolving \underline{\textbf{A}}gents through Experience-\underline{\textbf{G}}rounded \underline{\textbf{A}}bstraction), a framework that enables continual evolution of LLM agents through experience-grounded knowledge abstraction. 
SAGA establishes a closed-loop interaction between task execution and knowledge evolution, where acquired experiences are transformed into reusable knowledge and subsequently leveraged to support future decision making.

As illustrated in Fig.~\ref{fig:framework}, SAGA integrates experience abstraction with knowledge-guided execution in a self-evolving loop. 
During task execution, the agent collects interaction trajectories from the environment. 
These trajectories are progressively abstracted into multiple levels of knowledge, including meta descriptions, reusable strategies, and experience-grounded principles, which are organized in a memory library. 
For subsequent tasks, relevant knowledge is retrieved to provide decision guidance, while extracted principles further serve as high-level behavioral constraints for regulating generated actions. 
If a candidate action is inconsistent with the acquired principles, SAGA refines the action through resampling. 
The resulting interactions generate new experiences that further update and refine the knowledge representations, enabling continual agent evolution.

\subsection{Hierarchical Knowledge Representation}\label{sec:knowledge_representation}

To enable experience-grounded knowledge evolution, SAGA organizes agent memory
into multiple abstraction levels. 
Following the experience compression perspective, where memories, skills, and
rules represent progressively abstract knowledge representations
~\citep{zhang2026experience}, SAGA transforms concrete interactions into
reusable knowledge while preserving the evidence chain from execution
trajectories. 
Specifically, SAGA considers four levels of abstraction: raw experiences,
episodic knowledge, procedural knowledge, and experience-grounded principles.

\textbf{$L_0$: Raw Experience.}
Raw experiences represent the original interaction trajectories collected
during task execution. 
A trajectory is represented as
\[
\tau_i=\left(s_0,\{(a_t,o_t,r_t)\}_{t=1}^{T}\right),
\]
where $o_t$, $a_t$, and $r_t$ denote the observation, action, and environmental
feedback at step $t$, respectively. 
Raw experiences preserve fine-grained execution details and serve as the
evidence source for subsequent abstraction. 
However, their strong dependency on individual task instances limits
transferability across diverse tasks and environments.

\begin{figure}[t]
\centering
\begin{tikzpicture}[
    box/.style={
        draw,
        rounded corners,
        align=center,
        font=\scriptsize,
        minimum width=2.9cm,
        minimum height=0.7cm
    },
    l0/.style={box, fill=myblue!20, draw=myblue},
    l1/.style={box, fill=mygreen!20, draw=mygreen},
    l2/.style={box, fill=myyellow!25, draw=myyellow},
    l3/.style={box, fill=myred!20, draw=myred},
    arrow/.style={->, thick}
]

\node[l3] (l3) {
Condition: Primary pigment mixing\\
Effect: Secondary color generation\\
Mechanism: Subtractive mixing rule
};

\node[l2, below left=0.4cm and 0.6cm of l3] (l2a) {
Pattern: Color mixing task\\
Strategy: Select pigments,\\
combine and mix
};

\node[l2, below right=0.4cm and -0.8cm of l3] (l2b) {
Pattern: Color generation\\
Strategy: Mix pigments\\
and verify outcome
};

\node[l1, below left=0.5cm and 0.6cm of l2a] (l1a) {
Description: Mixed yellow and blue\\
pigments to generate green pigment\\
Result: Successful execution
};

\node[l1, below=0.5cm of l2a] (l1b) {
Description: $\dots$\\
Result: $\dots$
};

\node[l1, below right=0.5cm and 0.6cm of l2a] (l1c) {
Description: Completed a pigment mixing\\
task with the desired color obtained\\
Result: Successful execution
};

\node[l1, below=1.9cm of l2b] (l1d) {
$\dots$
};

\node[l0, below=0.5cm of l1b] (l0) {
Pour yellow pigment\\
Mix (insufficient)\\
Pour blue pigment\\
Mix $\rightarrow$ green pigment
};

\draw[arrow] (l0) -- (l1b);

\draw[arrow] (l1a) -- (l2a);
\draw[arrow] (l1b) -- (l2a);
\draw[arrow] (l1c) -- (l2a);
\draw[arrow] (l1d) -- (l2b);

\draw[arrow] (l2a) -- (l3);
\draw[arrow] (l2b) -- (l3);

\end{tikzpicture}

\caption{Illustration of hierarchical knowledge abstraction in SAGA. 
The colors represent different abstraction levels: 
\textcolor{myblue}{blue} for $L_0$ raw experiences, 
\textcolor{mygreen}{green} for $L_1$ episodic knowledge, 
\textcolor{myyellow}{yellow} for $L_2$ procedural knowledge, and 
\textcolor{myred}{red} for $L_3$ experience-grounded principles.}
\vspace{-1.5em}
\label{fig:hierarchy_example}
\end{figure}
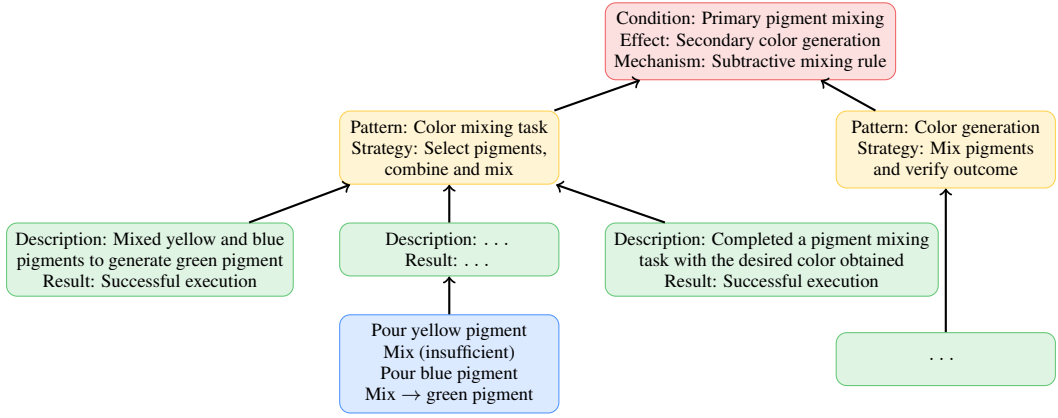

\textbf{$L_1$: Episodic Knowledge.}
Episodic knowledge provides a compact representation of individual interaction
experiences. 
Each episodic representation preserves task-relevant information, including
task objectives, key interactions, outcomes, and encountered obstacles, while
removing redundant execution details. 
Given a raw trajectory $\tau_i$, an LLM-based summarization process generates
its corresponding episodic representation:
\[
\mathcal{K}^{1}_i=\mathrm{Summarize}_{\mathrm{LLM}}(\tau_i).
\]
Thus, $L_0$ and $L_1$ maintain a one-to-one correspondence, where each episodic
representation is grounded in a specific trajectory while providing a compact
representation for retrieval and further abstraction.

\textbf{$L_2$: Procedural Knowledge.}
Procedural knowledge captures reusable solution patterns aggregated from
multiple episodic experiences with shared task structures. 
Unlike episodic knowledge that represents individual executions, procedural
knowledge describes generalized strategies for a class of related tasks.

Specifically, SAGA identifies related episodic representations and extracts
procedural knowledge through an LLM-based extraction process:
\[
\mathcal{C}^{1}_i=\{\mathcal{K}^{1}_{j}\}_{j\in \mathcal{I}_i},\qquad
\mathcal{K}^{2}_i=
\mathrm{Extract}_{\mathrm{LLM}}
\left(\mathcal{C}^{1}_i,\{\tau_j\}_{j\in\mathcal{I}_i}\right),
\]
where $\mathcal{I}_i$ denotes the index set of episodic experiences sharing a
common task pattern. 
The extracted procedure summarizes common behavioral structures across
experiences while preserving grounding in the original trajectories.
Multiple episodic experiences may contribute to one procedural representation,
and different procedures may share overlapping experiences.

\textbf{$L_3$: Experience-Grounded Principles.}
Experience-grounded principles capture fundamental regularities inferred from
multiple procedural representations and their supporting experiences. 
They describe underlying rules, constraints, and relations shared across
different procedures.

Specifically, SAGA infers principle knowledge from related procedures through
an LLM-based inference process:
\[
\mathcal{C}^{2}_i=\{\mathcal{K}^{2}_{j}\}_{j\in \mathcal{J}_i},\qquad
\mathcal{K}^{3}_i=
\mathrm{Infer}_{\mathrm{LLM}}
\left(\mathcal{C}^{2}_i,\mathcal{E}_i\right),
\]
where $\mathcal{J}_i$ denotes the index set of related procedures, and
$\mathcal{E}_i$ denotes their supporting evidence. 
The resulting principles provide higher-level guidance beyond individual
procedures while maintaining traceability to underlying experiences.

Figure~\ref{fig:hierarchy_example} illustrates this abstraction process using
a pigment mixing example from ScienceWorld, where multiple interaction
experiences are progressively transformed into reusable procedures and
higher-level principles stored in the hierarchical memory $\mathcal{M}$.

\subsection{Knowledge Utilization and Self-Evolving Loop}

\begin{algorithm}[t]
\caption{Knowledge Utilization and Self-Evolving Loop in SAGA}
\label{alg:saga}
\begin{algorithmic}[1]

\REQUIRE Task instance $x_t$, hierarchical memory $\mathcal{M}$

\STATE Generate task-level plan:
$p_t\leftarrow\mathrm{Plan}_{\mathrm{LLM}}(x_t)$

\STATE Retrieve hierarchical knowledge:
$\mathcal{Z}_t\leftarrow\mathrm{Retrieve}(\mathcal{M},x_t,p_t)$

\STATE Instantiate retrieved principles:
$\tilde{\mathcal{Z}}^{\mathrm{prin}}_t
\leftarrow
\mathrm{Instantiate}(\mathcal{Z}^{\mathrm{prin}}_t,x_t,p_t)$

\STATE Initialize trajectory $\tau_t\leftarrow\emptyset$

\FOR{each decision step $k$}

    \STATE Generate candidate action:
    $a_k\sim\pi(\cdot|o_k,\tilde{\mathcal{Z}}_t)$

    \STATE Evaluate action using principle predicates:
    $\{\phi_i(a_k)\}_{i=1}^{N_t}$

    \IF{$\exists \phi_i,\phi_i(a_k)=0$}

        \STATE Provide violated principle information for action refinement

        \STATE Regenerate action

    \ENDIF

    \STATE Execute action $a_k$ and collect feedback $r_k$

    \STATE Update trajectory:
    $\tau_t\leftarrow\tau_t\cup(o_k,a_k,r_k)$

\ENDFOR

\STATE Abstract trajectory following hierarchical abstraction:
$\tau_t\rightarrow\mathcal{K}_t$

\STATE Update memory:
$\mathcal{M}\leftarrow\mathrm{Update}(\mathcal{M},\mathcal{K}_t)$

\RETURN Updated memory $\mathcal{M}$

\end{algorithmic}
\end{algorithm}

Given a task instance $x_t$ and the hierarchical memory $\mathcal{M}$,
SAGA performs knowledge-guided execution and self-evolution through an
iterative retrieval--utilization--abstraction loop, as summarized in
Algorithm~\ref{alg:saga}. 
Different from conventional retrieval-based agents that directly append
retrieved information into the context, SAGA explicitly adapts retrieved
knowledge to the current task and incorporates it into both decision making
and knowledge evolution.

The process begins with task-aware knowledge retrieval. 
The agent first generates a task-level plan
$p_t=\mathrm{Plan}_{\mathrm{LLM}}(x_t)$, which provides additional semantic
information about the task objective, intermediate requirements, and potential
execution structure. 
The task description and the generated plan are then used to retrieve relevant
knowledge from the hierarchical memory:
\[
\mathcal{Z}_t=
\mathrm{Retrieve}(\mathcal{M},x_t,p_t),
\]
where $\mathcal{Z}_t$ contains task-relevant knowledge from different
abstraction levels introduced in Section~\ref{sec:knowledge_representation}.
To enable effective utilization of high-level knowledge, SAGA performs
task-conditioned principle instantiation:
\[
\tilde{\mathcal{Z}}^{\mathrm{prin}}_t
=
\mathrm{Instantiate}
(\mathcal{Z}^{\mathrm{prin}}_t,x_t,p_t).
\]
This step grounds abstract experience-derived principles into task-specific
guidance by incorporating the current objective and environmental context.
The resulting knowledge context $\tilde{\mathcal{Z}}_t$ is then provided to the
decision module for subsequent execution.

During execution, SAGA integrates retrieved knowledge into policy generation
and principle-level behavioral regulation. 
At each decision step $k$, the policy generates a candidate action
$a_k\sim\pi(\cdot|o_k,\tilde{\mathcal{Z}}_t)$ based on the current observation
and instantiated knowledge context. 
To make high-level principles executable, SAGA compiles the instantiated
$L_3$ experience-grounded principles into executable predicates:
\[
\mathcal{P}_t=\{\phi_i\}_{i=1}^{N_t},\qquad
\phi_i(a_k)\in\{0,1\},
\]
where each predicate evaluates whether the generated action satisfies the
corresponding principle constraint. 
The candidate action is accepted when all applicable predicates are satisfied.
When a violation is detected, SAGA identifies the violated principles and
provides their semantic information as feedback for action refinement.
Specifically, the violated principles are incorporated into the subsequent
decision context:
$\tilde{o}_k=(o_k,\{\phi_i|\phi_i(a_k)=0\})$,
which guides the policy to resample an alternative action that better aligns
with experience-grounded constraints.
This mechanism enables abstract principles to regulate low-level decisions
without replacing the underlying generative policy, achieving a balance
between knowledge-based guidance and behavioral flexibility.

After task completion, the collected trajectory
$\tau_t=\{(o_k,a_k,r_k)\}_{k=1}^{T}$ is incorporated into the knowledge
evolution process. 
Instead of storing the trajectory as an isolated experience, SAGA applies the
hierarchical abstraction mechanism introduced in
Section~\ref{sec:knowledge_representation} to extract reusable knowledge:
$\tau_t\rightarrow\mathcal{K}_t$.
The extracted knowledge is subsequently integrated into memory:
\[
\mathcal{M}_{t+1}
=
\mathrm{Update}(\mathcal{M}_t,\mathcal{K}_t).
\]
Successful executions provide additional evidence for reinforcing effective
strategies, while failed interactions reveal ineffective behaviors and support
knowledge refinement.
Through repeated retrieval, execution, and abstraction, SAGA forms a
self-evolving knowledge loop, where accumulated experiences are continuously
converted into reusable knowledge to improve future task adaptation.

\section{Experiments}
\subsection{Experimental Setup}

We evaluate SAGA on two interactive benchmarks, ScienceWorld~\citep{wang2022scienceworld} and ALFWorld~\citep{shridhar2020alfworld}, which require agents to perform long-horizon reasoning and decision making through continuous environment interaction. 
We consider two evaluation paradigms: \textbf{prompt-based experience evolution}, where the model parameters remain fixed and agent capabilities are improved through experience-grounded knowledge acquisition and utilization, and \textbf{training-based adaptation}, where accumulated experiences are further used to optimize model parameters. 
Unless otherwise specified, training-based methods ($\ast$) use Qwen2.5-7B-Instruct~\citep{qwen2025qwen25technicalreport} as the base model. 
For prompt-based methods ($\dagger$), we report results with the corresponding backbone models specified in each experiment.

We compare SAGA with representative baselines from three categories. 
The first category consists of reasoning-based agents, including ReAct~\citep{yao2022react}. 
The second category contains prompt-based experience augmentation approaches, including ExpeL~\citep{zhao2024expel}, Memento~\citep{zhou2025memento}, and ReasoningBank~\citep{ouyang2026reasoningbank}. 
The third category includes training-based adaptation methods, including reinforcement learning approaches such as GRPO~\citep{shao2024deepseekmath} and SkillRL~\citep{xia2026skillrl}, as well as recent self-evolving frameworks such as StraTA~\citep{xue2026strata} and Evolving-RL~\citep{fan2026evolving}. 
For ScienceWorld, we report the final task score averaged over evaluation instances, while for ALFWorld we follow the standard protocol and report task success rates.

\begin{table}[t]
\centering
\vspace{-1mm}
\caption{\small
Main results on ScienceWorld.
$\dagger$ denotes prompt-based methods and $\ast$ denotes training-based methods.
Best and second-best performances are highlighted in \textbf{bold} and \underline{underlined}, respectively.
}
\label{tab:scienceworld_main}
\scalebox{0.78}{
\begin{tabular}{llcccccc}
\toprule
Adaptation Paradigm
& Method 
& Measure 
& Test-Cond 
& Find 
& Chem-Mix 
& Lifespan 
& Avg. $\uparrow$\\
\midrule

\multirow{7}{*}{Prompt-based}
& GPT-5.1$^\dagger$
& 32.1 & 46.5 & 41.8 & \underline{69.6} & 38.3 & 43.0\\

& Claude-4-Sonnet$^\dagger$
& \underline{45.4} & 54.7 & 60.3 & 45.6 & \underline{90.0} & \underline{57.4}\\

& Gemini-2.5-Flash$^\dagger$
& 40.7 & 51.3 & 53.7 & 33.0 & 50.0 & 49.6\\

& Qwen2.5-7B-Instruct$^\dagger$
& 2.1 & 12.3 & 21.2 & 9.6 & 25.0 & 13.9\\

& ReAct$^\dagger$
& 12.2 & 14.4 & 33.6 & 9.6 & 28.9 & 20.0\\

& Ours (DeepSeek-flash)$^\dagger$
& 25.1 & \underline{58.1} & \underline{67.7} & 24.6 & 80.0 & 56.2\\

& Ours (Qwen3.8-Max)$^\dagger$
& \textbf{55.9} & \textbf{62.2} & \textbf{68.0} & \textbf{100.0} & \textbf{100.0} & \textbf{67.0}\\

\midrule

\multirow{7}{*}{Training-based}
& AgentGym-RL$^\ast$
& 24.5 & \underline{59.0} & 65.9 & 0.0 & 66.7 & 50.5\\

& ScalingInter$^\ast$
& 34.0 & 55.4 & \textbf{88.6} & 0.0 & 73.3 & 57.0\\

& PPO$^\ast$
& 1.8 & 57.5 & 60.9 & 6.9 & 97.8 & 51.4\\

& GRPO$^\ast$
& 31.2 & 49.8 & 32.7 & 14.7 & 53.3 & 41.8\\

& StraTA$^\ast$
& \textbf{55.7} & 54.4 & \underline{78.6} & 16.3 & \textbf{100.0} & \underline{63.5}\\

& Ours (SFT)$^\ast$
& 26.8 & 59.1 & 64.1 & \textbf{86.6} & 83.3 & 58.1\\

& Ours (SFT+GRPO)$^\ast$
& \underline{46.2} & \textbf{63.6} & 74.5 & \underline{69.6} & \textbf{100.0} & \textbf{66.9}\\

\bottomrule
\end{tabular}
}

\vspace{1mm}
\parbox{0.95\linewidth}{
\footnotesize
\textit{Note:}
Methods are grouped by adaptation paradigm.
For training-based variants, $L0$--$L3$ knowledge is incorporated into the model prompt during training.
}
\end{table}

\subsection{Main Results}

\textbf{ScienceWorld Results.} ScienceWorld evaluates an agent's capability to perform long-horizon scientific reasoning and procedural manipulation through interactive environments. 
We investigate whether experience-grounded knowledge evolution enables agents to acquire and reuse transferable knowledge. 
Table~\ref{tab:scienceworld_main} summarizes the results under both prompt-based and training-based adaptation paradigms. 
For prompt-based methods, SAGA consistently improves over conventional prompting and direct trajectory retrieval. 
Using DeepSeek-flash as the backbone, SAGA improves the average score from 25.0\% without memory and 54.8\% with trajectory retrieval to 56.2\% through hierarchical knowledge evolution, demonstrating the effectiveness of progressively abstracting concrete experiences into reusable knowledge representations. 
Furthermore, with the stronger Qwen3.8-Max backbone, SAGA achieves an average score of 67.0\%, indicating that the proposed experience-grounded abstraction framework generalizes across different backbone models. 
For training-based adaptation, SAGA further exploits accumulated experiences for parameter optimization. 
Starting from the Qwen2.5-7B-Instruct backbone (9.6\%, get by ourself), supervised fine-tuning improves the average score to 58.1\%, and subsequent GRPO optimization further increases the performance to 66.9\%, achieving the best overall performance among all compared methods. 
These results demonstrate that experience-grounded knowledge evolution provides a unified framework for improving LLM agents through both inference-time knowledge utilization and parameter-level adaptation.

\begin{figure}[t]
\centering
\begin{minipage}{0.52\linewidth}
\vspace{0pt}
\textbf{Evolution trajectory during GRPO optimization.}
To further illustrate how accumulated experience contributes to continual agent improvement, we visualize the performance trajectory during GRPO optimization in Fig.~\ref{fig:grpo_evolution}. 
The selected checkpoint improves the final evaluation performance from the SFT initialization to 66.9\%. 
Although intermediate updates exhibit fluctuations due to the stochastic nature of reinforcement learning optimization, the overall trend demonstrates continuous refinement of agent capabilities through experience-guided parameter adaptation.
\end{minipage}
\hfill
\begin{minipage}{0.43\linewidth}
\vspace{0pt}
\centering
\includegraphics[
    width=\linewidth,
    trim={140 240 120 60},
    clip
]{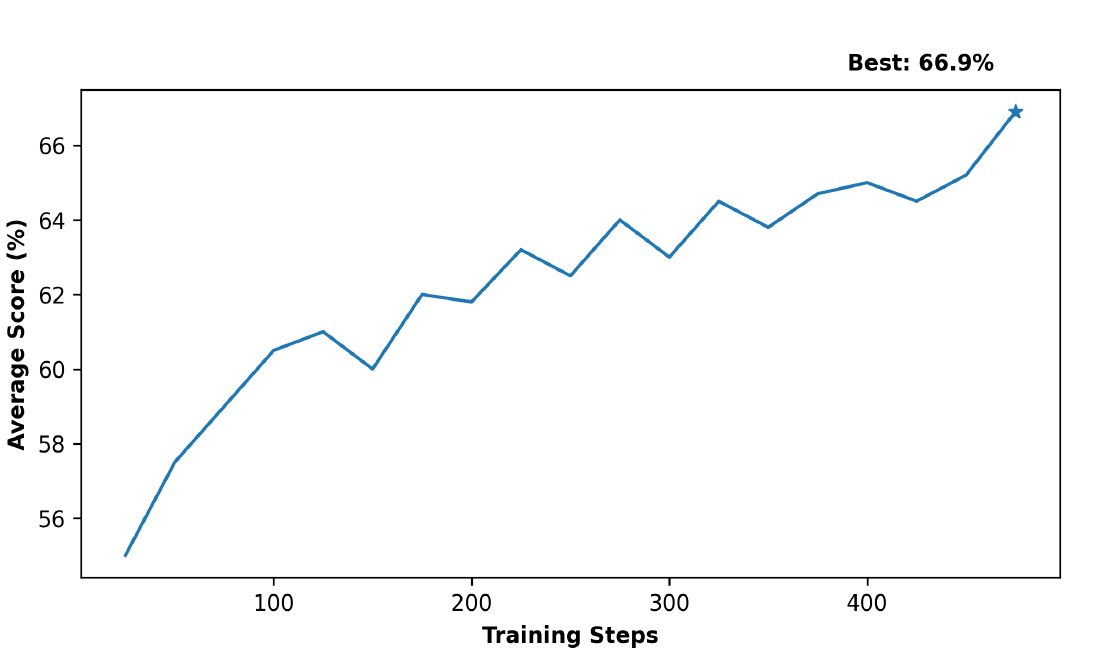}
\caption{Performance evolution during GRPO optimization in ScienceWorld.}
\label{fig:grpo_evolution}
\end{minipage}
\end{figure}

\begin{table}[t]
\centering
\vspace{-3mm}
\caption{\small
Main results on ALFWorld.
$\dagger$ denotes prompt-based methods and $\ast$ denotes training-based methods.
Best performances on the overall success rate and unseen tasks (Look/Pick2) are highlighted in \textbf{bold}.
}
\label{tab:alfworld_main}
\scalebox{0.78}{
\begin{tabular}{llccccccc}
\toprule
Adaptation Paradigm
& Method
& Pick
& Clean
& Heat
& Cool
& Look
& Pick2
& Avg. $\uparrow$\\
\midrule

\multirow{6}{*}{Prompt-based}
& Base Model$^\dagger$
&75.0&57.4&13.0&60.0&18.9&36.5&45.5\\

& ReasoningBank$^\dagger$
&75.0&45.2&21.7&47.6&11.1&35.3&41.1\\

& ExpeL$^\dagger$
&21.0&55.0&52.0&71.0&67.0&6.0&46.3\\

& Memento$^\dagger$
&79.2&64.5&39.1&76.2&0.0&41.2&53.0\\

& Ours (Qwen3.8-Max, no memory)$^\dagger$
&100.0&96.8&87.0&95.2&\textbf{100.0}&\textbf{100.0}&96.3\\

& Ours (Qwen3.8-Max, SAGA)$^\dagger$
&100.0&100.0&100.0&100.0&\textbf{100.0}&94.1&
\textbf{99.3}\\

\midrule

\multirow{8}{*}{Training-based}
& GRPO$^\ast$
&96.7&90.3&100.0&100.0&53.3&12.9&79.9\\

& GRPO (w/ skills)$^\ast$
&95.8&96.1&98.3&98.1&61.1&27.1&83.3\\

& SkillRL$^\ast$
&93.3&95.2&72.2&80.0&81.2&55.6&81.7\\

& Evolving-RL (w/o skills)$^\ast$
&94.2&96.8&99.1&100.0&\textbf{100.0}&61.2&93.1\\

& Evolving-RL (w/ skills)$^\ast$
&97.5&97.4&100.0&100.0&95.6&\textbf{81.2}&96.0\\

& Ours (SFT)$^\ast$
&100.0&100.0&87.0&95.2&88.9&58.8&90.3\\

& Ours (SFT + GRPO)$^\ast$
&95.8&100.0&91.3&95.2&94.4&76.5&93.3\\

& Ours (SFT + GRPO + External)$^\ast$
&100.0&100.0&95.7&100.0&\textbf{100.0}&76.5&
\textbf{96.3}\\

\bottomrule
\end{tabular}
}
\vspace{1mm}
\parbox{0.95\linewidth}{
\footnotesize
\textit{Note:} Results with ``skills'' or ``SFT + GRPO + External'' is additionally allowed to collect skills from the unseen training subset.
}
\vspace{-2.2em}
\end{table}
\textbf{ALFWorld Results.} ALFWorld evaluates an agent's capability to execute long-horizon embodied instructions in interactive household environments. 
Table~\ref{tab:alfworld_main} summarizes the performance under both prompt-based and training-based adaptation paradigms. 
For prompt-based methods, experience-grounded knowledge evolution consistently improves agent performance by transforming interaction trajectories into reusable knowledge representations. 
With the Qwen2.5-7B-Instruct backbone, memory augmentation improves the success rate from 45.5\% to 57.5\%, demonstrating the effectiveness of structured experience utilization over direct prompting. 
Furthermore, for the stronger Qwen3.8-Max backbone, SAGA achieves 99.3\% overall success rate, indicating that hierarchical knowledge evolution can effectively complement the capability of stronger foundation models.

For training-based adaptation, SAGA further incorporates accumulated experiences into parameter optimization. 
Starting from the SFT-based model, SAGA achieves a success rate of 90.3\%, while subsequent GRPO optimization further improves the performance to 93.3\%. 
With additional external test-time adaptation, the performance increases to 96.3\%, achieving competitive performance with methods that explicitly leverage additional skills. 
In particular, SAGA maintains strong performance on unseen tasks, demonstrating the potential of experience-grounded knowledge evolution for improving generalization in long-horizon embodied decision making.

\subsection{Ablation Studies}

We conduct ablation studies on ScienceWorld to examine three
components of SAGA: hierarchical knowledge abstraction,
task-conditioned principle instantiation, and principle-based
action regulation.
Table~\ref{tab:saga_ablation} compares six configurations using
Qwen3.8-Max and DeepSeek-v4-Flash, progressively introducing
different knowledge representations and utilization mechanisms.

\textbf{Effect of hierarchical knowledge abstraction.}
Retrieving raw trajectories (R2) substantially improves performance
over memory-free execution (R1), increasing the score from
48.4 to 63.7 with Qwen3.8-Max and from 25.0 to 54.8 with
DeepSeek-v4-Flash.
These improvements establish direct experience reuse as a strong
reference configuration.
Replacing raw trajectories with episodic and procedural knowledge
(R3) yields further gains of 0.8 and 0.2 points, respectively.
However, directly adding principle-level knowledge (R4) decreases
the score relative to R3 for both backbones.
This pattern indicates that more abstract knowledge does not
automatically translate into better execution, motivating
mechanisms that explicitly adapt principles to the current task.

\textbf{Effect of contextual instantiation.}
The comparison between R4 and R5 examines the contribution of
task-conditioned principle instantiation.
Converting retrieved principles into operational guidance improves
the score from 63.9 to 66.2 with Qwen3.8-Max and from 54.5 to
55.8 with DeepSeek-v4-Flash.
Moreover, R5 exceeds the episodic-plus-procedural configuration
(R3) by 1.7 and 0.8 points.
Thus, principle-level knowledge provides additional value when
its applicability and operational implications are grounded in
the current task.
The consistent ordering across both backbones supports the
importance of contextual instantiation for effective utilization
of abstract knowledge.

\textbf{Effect of principle-based action regulation.}
Adding action regulation to instantiated guidance (R6) further
increases the score to 67.0 with Qwen3.8-Max and 56.2 with
DeepSeek-v4-Flash, corresponding to gains of 0.8 and 0.4 points
over R5.
These incremental improvements are consistent with the role of
principle-conditioned feedback in refining candidate actions
during execution.
The complete configuration also outperforms raw trajectory
retrieval by 3.3 and 1.4 points, respectively.
Together, the comparisons support a complementary role for
contextual instantiation and action regulation in converting
experience-derived principles into effective decision guidance.

\begin{table}[t]
\centering
\vspace{-1mm}
\caption{\small
Ablation study on knowledge abstraction and utilization on ScienceWorld.
}
\label{tab:saga_ablation}

\scalebox{0.82}{
\begin{tabular}{lcc}
\toprule
Configuration
& Qwen3.8-Max
& DeepSeek-v4-Flash \\
\midrule

R1: No memory
& 48.4
& 25.0 \\

R2: Raw trajectory retrieval ($L_0$)
& 63.7
& 54.8 \\

R3: Episodic + procedural knowledge ($L_1+L_2$)
& 64.5
& 55.0 \\

R4: $L_1+L_2+L_3$, direct injection
& 63.9
& 54.5 \\

R5: $L_1+L_2$ + instantiated $L_3$
& 66.2
& 55.8 \\

R6: R5 + principle-based action regulation
& \textbf{67.0}
& \textbf{56.2}\\

\bottomrule
\end{tabular}
}
\vspace{-1.6em}
\end{table}

\section{Conclusion}

In this work, we study continual agent evolution through experience-grounded knowledge abstraction and introduce SAGA, a framework that converts interaction trajectories into progressively more reusable forms of knowledge. By organizing experience into episodic descriptions, reusable procedures, and principles with explicit applicability conditions, SAGA enables agents to move beyond task-specific memory while preserving links between abstract knowledge and its supporting execution evidence. SAGA further closes the loop between knowledge acquisition and action by instantiating retrieved principles into task-specific guidance and using principle-derived checks to regulate subsequent decisions. Experiments on ScienceWorld and ALFWorld show consistent performance improvements, while ablations confirm the importance of both contextual instantiation and action regulation for realizing the benefits of principle-level knowledge. These results suggest that continual improvement in LLM agents depends not only on retaining past experience, but also on systematically abstracting, grounding, and operationalizing that experience for future interaction.

\bibliography{iclr2027_conference}
\bibliographystyle{iclr2027_conference}

\newpage
\appendix

\section{Hierarchical Memory Construction}
\label{app:memory}

This section formalizes the persistent representation, hierarchical extraction pipeline, and incremental update operators of memory $\mathcal{M}$, complementing the framework-level specification in Algorithm~\ref{alg:saga}.

\noindent\textbf{Schema and persistence.}
Each level of $\mathcal{M}$ is maintained as an append-only store with an associated embedding index.
An $L_0$ node preserves the raw trajectory $\tau$, including stepwise observations, actions, environment feedback, and the terminal outcome returned by the environment; this authoritative outcome is never replaced by a model-generated summary.
An $L_1$ node encodes the decontextualized goal, key actions, outcome, and obstacle associated with one trajectory, thereby preserving a one-to-one correspondence with its $L_0$ evidence.
An $L_2$ node represents a transferable behavioral pattern and an ordered, multi-step strategy supported by multiple episodic records rather than by an isolated action.
An $L_3$ node specifies a condition, effect, mechanism, scope, and set of exceptions. Such nodes are interpreted as evidence-grounded explanatory hypotheses, rather than as verified causal laws.
Bidirectional parent-child links connect adjacent levels, with each lower-level node assigned to at most two parents to preserve a sparse and auditable evidence graph.
Source metadata records the environment, task, variant, split, and round, providing end-to-end traceability from each principle to its supporting trajectories.
Retrieval uses Qwen3-0.6B-embedding with cosine similarity. The indexed representation comprises the goal field at $L_1$, the pattern and strategy fields at $L_2$, and the scope, condition, effect, and mechanism fields at $L_3$. Because the index is fully reconstructible from the persistent stores, it does not constitute an independent knowledge source.

\noindent\textbf{Cold start and hierarchical extraction.}
Memory initialization relies exclusively on memory-free interactions with training tasks and introduces no human-authored seed knowledge.
At each subsequent round, the actor interacts with the environment under the previously frozen memory state, collects additional training trajectories, and incrementally updates the inherited store.
An LLM transforms each eligible raw trajectory into an $L_1$ record while preserving distinctions among successful outcomes, task failures, and encountered obstacles. Episodes containing no executed action are excluded, whereas API failures are recorded separately from environment-level failures.
For $L_2$ induction, records are grouped by task and outcome, and each group is consolidated into a compact set of complete procedures with explicit citations to supporting cases. Lexical similarity alone is not treated as evidential support.
For $L_3$ construction, the builder jointly considers related procedures and their supporting evidence to infer a shared condition, effect, mechanism, scope, and set of exceptions. Procedures for which this abstraction is insufficiently supported remain at $L_2$.
Cross-task merging uses semantic similarity with a threshold of $0.75$ to generate candidate procedure clusters. A merged principle is admitted only when its evidence spans at least two task types; similarity is used for candidate generation, not as evidence of a shared mechanism.

\noindent\textbf{Incremental update.}
Algorithm~\ref{alg:update} specifies the update operator applied to the current memory and newly collected training trajectories.
The selection stage retains at most one trajectory per task variant, ranked lexicographically by terminal score, peak score, and recency. Ties are resolved in favor of the more recent round without removing provenance, and trajectories rejected at this stage do not participate in the current abstraction update.
ScienceWorld admits trajectories with a positive peak score. ALFWorld additionally admits executed failures, including zero-score episodes, as negative evidence for actions or strategies that should be avoided.
Each newly induced procedure is first matched to existing principles for the same task. Only unmatched procedures are considered for principle lifting, reducing redundant abstraction.
Candidate principles are deduplicated, and semantically equivalent candidates are consolidated in proportion to the support supplied by their underlying evidence.
Principles that fail validation are recorded in an exclusion registry, preventing semantically equivalent candidates from being reintroduced in subsequent rounds.
After each update, the resulting memory is frozen for evaluation. Evaluation trajectories are excluded from all subsequent construction stages.
For each round, we record trajectory admissions, replacements, and rejections; $L_2$ and $L_3$ counts; principle reuse; cross-task evidence coverage; validation outcomes; and model cost. These structural statistics characterize memory evolution but do not independently establish performance improvement.

\begin{algorithm}[t]
\caption{Incremental Memory Update}
\label{alg:update}
\begin{algorithmic}[1]
\REQUIRE Current memory $\mathcal{M}=\{L_0,L_1,L_2,L_3\}$; new training trajectories $\mathcal{T}$; exclusion registry $\mathcal{R}$; merge threshold $\theta{=}0.75$
\STATE Initialize selected set $\mathcal{T}^{\star}\leftarrow\emptyset$
\FOR{each task variant $v$}
    \STATE Rank trajectories of $v$ by terminal score, then peak score, then recency
    \STATE Add the top trajectory admitted by the benchmark rule to $\mathcal{T}^{\star}$
\ENDFOR
\STATE Append $\mathcal{T}^{\star}$ to $L_0$ and summarize each trajectory into an episodic record in $L_1$
\STATE Group the new $L_1$ records by task and outcome; induce candidate procedures with supporting-case citations
\FOR{each candidate procedure $c$}
    \IF{$c$ matches an existing same-task principle}
        \STATE Attach $c$ as supporting evidence without lifting a new principle
    \ELSE
        \STATE Lift $c$ to a candidate principle
    \ENDIF
\ENDFOR
\STATE Deduplicate candidates; merge a pair with similarity ${\ge}\theta$ only when its evidence spans ${\ge}2$ task types
\STATE Remove candidates whose normalized form already appears in $\mathcal{R}$
\STATE Validate survivors (ALFWorld: evidence grounding; ScienceWorld: utility ledger)
\STATE Commit validated procedures and principles to $L_2$ and $L_3$; append rejected candidates to $\mathcal{R}$
\STATE Freeze the updated memory $\mathcal{M}'$ and exclude evaluation trajectories from future construction
\RETURN Updated memory $\mathcal{M}'$
\end{algorithmic}
\end{algorithm}

\noindent
The benchmark admission rule in the selection stage retains trajectories with a positive peak score for ScienceWorld, whereas for ALFWorld it additionally retains executed failures, including zero-score episodes, as negative evidence.

\noindent\textbf{Evidence validation.}
The two benchmarks employ benchmark-specific quality-control procedures targeting complementary notions of validity.
For ALFWorld, an evidence-grounding checker verifies citation syntax, source integrity, realized parent--child links, and consistency between each candidate principle and its supporting trajectories. A cross-task principle must be supported by evidence from at least two task types; textual expansion of its stated scope is insufficient. An API fault prevents the affected stage from being committed and is counted as neither a candidate rejection nor an environment-level task failure.
For ScienceWorld, a utility ledger compares paired rollouts with and without knowledge injection on training variants, indexed by task and recalled set. The ledger supports collection screening, leave-one-out diagnosis, and pruning, but applies only to experiments in which utility screening is explicitly enabled; earlier batches are not retrospectively classified as ledger-validated.
Evidence grounding evaluates whether cited trajectories support a principle, whereas the utility ledger estimates the empirical effect of deploying that principle. These criteria are therefore reported separately.

\noindent\textbf{Prompt specification.}
Table~\ref{tab:prompt_spec} summarizes the input-output contract of each LLM-based component.
Complete prompt templates, implementation defaults, and entry-point arguments are provided in the anonymized supplementary code.

\begin{table}[t]
\centering
\vspace{-1mm}
\caption{\small
Input-output specification of the LLM-based components in SAGA.
}
\label{tab:prompt_spec}
\scalebox{0.86}{
\begin{tabular}{lll}
\toprule
Component & Input & Output \\
\midrule
Planning & Task description & One-sentence strategy with a scalar confidence \\
$L_1$ summarization & Raw trajectory & Goal, key actions, outcome, obstacle \\
$L_2$ induction & Grouped $L_1$ records with trajectories & Pattern, strategy, supporting-case citations \\
$L_3$ lifting & Related $L_2$ procedures with evidence & Condition, effect, mechanism, scope, exceptions \\
Cross-task merge & Candidate procedure clusters & Merged principle with multi-task evidence \\
Instantiation & Principle, task, scene hints & One operational constraint or \textsc{not applicable} \\
Predicate compilation & Principle & Executable check $\phi_{\mathcal{P}}$ for action regulation \\
\bottomrule
\end{tabular}}
\end{table}

\section{Knowledge Retrieval and Utilization}
\label{app:utilization}

\noindent\textbf{Planning and retrieval.}
Given a task instance $x_t$, the actor produces a one-sentence plan $p_t$ together with a scalar confidence estimate. The concatenation of $x_t$ and $p_t$ forms the retrieval query; the plan conditions retrieval without superseding task-specific evidence.
The default direct-retrieval path matches $L_1$ records by goal, $L_2$ procedures by pattern and strategy, and $L_3$ principles by scope, condition, effect, and mechanism, retaining the top $k{=}3$ items at each level.
Plan confidence determines the budget for principle-level knowledge. No principle is injected when confidence is at least $0.8$; at most one is injected for confidence in $[0.5,0.8)$; and at most three are injected otherwise. Unparseable confidence outputs are assigned the conservative maximum budget of three principles.
We retain two alternatives for ablation: an evidence path that aggregates scores upward through parent links and a chain path that propagates scores from $L_1$.
Each run records the allocated budget and the numbers of retrieved and injected items. The logs distinguish four outcomes that would otherwise be conflated in aggregate statistics: knowledge deemed unnecessary, not retrieved, filtered, or instantiated as \textsc{not applicable}.

\noindent\textbf{Instantiation and presentation.}
Each retrieved principle is instantiated once per episode. Conditioned on the principle's mechanism field, the current task, and available scene information, the instantiator produces a single operational constraint; the output \textsc{not applicable} suppresses injection of that principle.
The resulting context comprises three labeled blocks: \texttt{[Similar case]} supplies episodic evidence and its outcome, \texttt{[Procedure]} specifies the complete strategy, and \texttt{[Mechanism]} provides the instantiated constraint.
The accompanying instruction characterizes retrieved knowledge as applicability-dependent guidance rather than as a prescriptive action sequence. Explicit task instructions retain precedence over memory content.
During prompt-based execution, SAGA supplies the actor with retrieved $L_1$--$L_3$ knowledge while retaining $L_0$ as provenance. Raw $L_0$ trajectories are rendered only in the designated trajectory-retrieval control. The reported training-based variants use the corresponding $L_0$--$L_3$ context in their training prompts, as specified in the main tables.

\noindent\textbf{Principle-compiled action regulation.}
Principles identified as harmful are compiled into executable predicates $\mathcal{P}$. A candidate action $a_t$ is accepted by the regulator only if $\phi_{\mathcal{P}}(a_t){=}1$.
A predicate violation initiates bounded, principle-conditioned resampling for at most three attempts. If no sampled candidate satisfies the predicate, the actor executes the original action. This soft intervention preserves the support of the generative policy while supplying principle-level corrective feedback.
Successful compilation alone does not establish predicate validity. We therefore audit predicate polarity, empirical trigger frequency, and false-penalty risk separately; principles labeled as beneficial or neutral are never repurposed as negative constraints.

\section{Evaluation Protocols and Optimization Details}
\label{app:protocol}

The two benchmarks share the hierarchical memory representation and selected extraction components, while adopting benchmark-specific interaction protocols and evaluation metrics.
Tables~\ref{tab:sci_config} and~\ref{tab:alf_config} define the corresponding evaluation configurations; Table~\ref{tab:model_roles} identifies the actor and builder used in each reported setting.

\begin{table}[t]
\centering
\vspace{-1mm}
\caption{\small
ScienceWorld evaluation configuration (strict alignment with StraTA~\citep{xue2026strata}).
}
\label{tab:sci_config}
\scalebox{0.86}{
\begin{tabular}{ll}
\toprule
Parameter & Value \\
\midrule
Evaluation set & strata175: StraTA official test split (200 items) filtered by 14 available levels \\
Composition & 175 unique (task, variation) items, 11 tasks, 5 categories \\
Category weights & Measure 28, Test-Cond 83, Find 44, Chem-Mix 5, Lifespan 15 \\
Step limit & 20 \\
Interaction history & Most recent 2 steps, no length-based episode termination \\
Admissible-action list & Not provided \\
Metric & Terminal-step score, item mean $\times 100$ \\
Seed & 42 \\
\bottomrule
\end{tabular}}
\end{table}

\noindent\textbf{ScienceWorld.}
Following the reference evaluation harness, we report the terminal-step score rather than the maximum intermediate score.
All SAGA rows in Table~\ref{tab:scienceworld_main} use the strict configuration in Table~\ref{tab:sci_config}, matching StraTA in evaluation set, step limit, history window, and admissible-action exposure. Results for comparison methods are reproduced from the values reported by \citet{xue2026strata} under their stated protocol.
The no-memory Qwen2.5-7B-Instruct score of $13.9$ in Table~\ref{tab:scienceworld_main} is likewise taken from \citet{xue2026strata}. Under our strict implementation of Table~\ref{tab:sci_config}, the same backbone obtains $9.6$ without memory; this independently measured score serves as the initialization reference for the training-based experiments.
Prompt-based evolution comprises six rounds with DeepSeek-flash (deepseek-v4-flash) and seven rounds with Qwen3.8-Max (qwen3.8-max-0902); the reported DeepSeek-flash result uses the round-5 memory state. At each round, trajectories are collected exclusively from training tasks, the memory is updated and frozen, and the resulting state is subsequently evaluated on strata175.

\begin{table}[t]
\centering
\vspace{-1mm}
\caption{\small
ALFWorld evaluation configuration. Dataset composition and success accounting follow Evolving-RL~\citep{fan2026evolving}; the interaction protocol is reported separately for each model family.
}
\label{tab:alf_config}
\scalebox{0.82}{
\begin{tabular}{ll}
\toprule
Parameter & Value \\
\midrule
Evaluation set & \texttt{valid\_unseen}, 134 unique games \\
Task types & Pick 24, Clean 31, Heat 23, Cool 21, Look 18, Pick2 17 \\
Seen / unseen split & Four seen types (99 games); Look and Pick2 unseen (35 games) \\
Training pool & 2{,}432 games: Pick 790, Clean 650, Heat 459, Cool 533 \\
Step limit & 50 \\
Strict 7B protocol & Token-prefix continuation; 16{,}384-token context \\
Strong / training protocol & Multi-turn chat; 16{,}384-token base context \\
Extended-context rows & 32{,}768-token context, explicitly marked in the reported configuration \\
Decoding & At most 512 tokens per step; temperature 0, top-$p$ 1, seed 42 \\
Action list & Environment order; exact \texttt{help} excluded \\
Metric & Environment-reported success rate $\times 100$ \\
\bottomrule
\end{tabular}}
\end{table}

\noindent\textbf{ALFWorld.}
All evaluation scenes are drawn from \texttt{valid\_unseen}. Here, ``seen'' refers to task types represented in the training pool, rather than to evaluation scenes encountered previously. No evaluation scene contributes to trajectory collection or memory construction.
Look and Pick2 remain zero-shot in every non-External configuration: neither task type contributes training trajectories, construction evidence, or test-time adaptation data. The explicitly marked External variant relaxes only the prohibition on test-time adaptation.
The fixed-7B strict experiments follow the official solver protocol (reference commit \texttt{7ed3d5867b19dbf21a5ddccfea7b77dccfb5346f}; protocol \texttt{alf-official-token-v2}). The chat template is applied once, actions are generated by token-prefix completion, outputs ending with \texttt{finish\_reason=length} are not executed, and actions receive no automatic correction.
The Qwen3.8-Max and training-based experiments instead follow the separately specified multi-turn chat protocol. Consequently, comparisons across these protocol families are not interpreted as controlled estimates of model-only effects.
At every evolution round, trajectory collection is restricted to the training pool, and the updated memory is frozen before evaluation on all 134 games.
In the strict 7B experiments, the no-memory baseline uses online concurrency 32, whereas evolution uses online concurrency 8 and offline-builder concurrency 8. Because serving schedules may affect token-level execution, this difference is reported as a potential implementation-level confound.

\begin{table}[t]
\centering
\vspace{-1mm}
\caption{\small
Actor and builder assignments for the reported configurations. The actor denotes the execution policy or the policy being optimized; the builder constructs abstractions and knowledge-conditioned training context.
}
\label{tab:model_roles}
\scalebox{0.84}{
\begin{tabular}{lll}
\toprule
Configuration & Actor (execution) & Builder (abstraction) \\
\midrule
ScienceWorld, prompt-based & DeepSeek-flash & DeepSeek-flash \\
ScienceWorld, prompt-based & Qwen3.8-max-0902 & Qwen3.8-max-0902 \\
ScienceWorld, training-based & Qwen2.5-7B-Instruct & Qwen3.8-max-0902 (offline) \\
ALFWorld, prompt-based (strict) & Qwen2.5-7B-Instruct (fixed) & Qwen3.8-max-0902 (offline) \\
ALFWorld, prompt-based (strong) & Qwen3.8-max-0902 & Qwen3.8-max-0902 \\
ALFWorld, training-based & Qwen2.5-7B-Instruct & Qwen3.8-max-0902 (offline) \\
\bottomrule
\end{tabular}}
\end{table}

\noindent\textbf{Model roles and attribution boundaries.}
In both training-based settings, Qwen2.5-7B-Instruct is the optimized policy, whereas Qwen3.8-max-0902 acts as the offline builder of hierarchical abstractions and knowledge-conditioned training context.
The strict ALFWorld configuration applies the same division of labor while holding the online 7B actor fixed; we refer to this setting as API-assisted memory evolution.
Prompt constraints and evidence validation reduce unsupported abstractions but cannot isolate contributions from the builder's pretrained knowledge. The reported effects therefore characterize the complete actor-builder pipeline rather than a same-model abstraction control.

\noindent\textbf{External test-time adaptation.}
The SFT\,+\,GRPO\,+\,External row in Table~\ref{tab:alfworld_main} permits test-time skill acquisition on the two unseen task types, consistent with the evaluation boundary of Evolving-RL (w/ skills). We report this configuration separately because it is not a zero-shot result.

\noindent\textbf{Supervised fine-tuning.}
The supervised training pipeline explicitly separates trajectory collection, offline abstraction, and policy optimization.
For ScienceWorld, trajectories collected during prompt-based evolution are replayed to reconstruct stepwise observations and action targets. Qwen3.8-max-0902 supplies offline abstractions and knowledge context, whereas Qwen2.5-7B-Instruct is optimized as the execution policy.
The training corpus comprises three sample categories, summarized in Table~\ref{tab:sft_corpus}: stepwise actions from full-score trajectories (8{,}788 samples), prefixes of zero-final-score trajectories truncated at the first scoring peak (4{,}882 samples), and peak prefixes from partial-score trajectories (9{,}292 samples). Because these inclusion criteria depend on observed scores, they do not imply semantic optimality of every retained action.
Knowledge-conditioned examples append the retrieved $L_0$--$L_3$ context to their inputs, consistent with Table~\ref{tab:scienceworld_main}. The auxiliary instantiation objective maps a principle, task, and scene to an operational constraint.
For ALFWorld, the teacher corpus consists of successful Qwen3.8-Max trajectories from the four seen task types. The corpus is partitioned into training and held-out validation subsets for full-parameter fine-tuning of Qwen2.5-7B-Instruct. The reported SFT variant is knowledge-conditioned; neither Look nor Pick2 contributes supervised trajectories.

\begin{table}[t]
\centering
\vspace{-1mm}
\caption{\small
ScienceWorld supervised fine-tuning corpus. Samples are stepwise action targets reconstructed from replayed prompt-based evolution trajectories.
}
\label{tab:sft_corpus}
\scalebox{0.86}{
\begin{tabular}{lc}
\toprule
Trajectory source & Samples \\
\midrule
Full-score trajectories (stepwise actions) & 8{,}788 \\
Zero-final-score trajectories (prefix to first scoring peak) & 4{,}882 \\
Partial-score trajectories (peak prefix) & 9{,}292 \\
\midrule
Total & 22{,}962 \\
\bottomrule
\end{tabular}}
\end{table}

\noindent\textbf{GRPO optimization.}
Table~\ref{tab:optim_hparams} lists the optimization configuration for both training-based arms.
Starting from the SFT checkpoint, we optimize a LoRA adapter using group-relative policy optimization (GRPO). Each training item generates eight trajectories, and variable-length step samples are pooled within each group without alignment by step index.
The reported ScienceWorld configuration uses a dense reward defined as the per-step score increment, without future-return accumulation. The ALFWorld configuration uses shaped rewards derived from environment success feedback. Trajectory-level terminal-score sharing, rule-based penalties, and principle-predicate penalties are retained as control variants.
Step-level advantages are computed from length-normalized log-probabilities of generated tokens. In the dense-reward setting, rewards are normalized by their within-group root-mean-square without mean subtraction, thereby preserving zeros and reward signs. The trajectory-level control instead employs group-centered advantages.
Each sampling batch induces a single parameter update, with no PPO-style multi-epoch reuse. The KL penalty relative to the frozen SFT reference has coefficient $0.02$, advantages are clipped to magnitude $3$, the learning rate is $5\times10^{-6}$, the maximum sequence length is $16{,}384$ tokens, and evaluation occurs every $25$ updates.
The ScienceWorld arm uses a 1{,}890-item training pool, two epochs, and 474 updates with 30{,}240 sampled episodes. The ALFWorld arms draw only from the 2{,}432-game pool of four seen task types; Look and Pick2 remain absent from parameter optimization.

\begin{table}[t]
\centering
\vspace{-1mm}
\caption{\small
Optimization configuration for the training-based variants. GRPO initializes from the SFT checkpoint and optimizes a LoRA adapter; entries marked --- are not applicable to the corresponding arm.
}
\label{tab:optim_hparams}
\scalebox{0.82}{
\begin{tabular}{lll}
\toprule
Stage / Parameter & ScienceWorld & ALFWorld \\
\midrule
SFT scheme & Full-parameter & Full-parameter \\
Base model & Qwen2.5-7B-Instruct & Qwen2.5-7B-Instruct \\
\midrule
GRPO adapter & LoRA & LoRA \\
Group size (trajectories per item) & 8 & 8 \\
Reward & Dense per-step score change & Shaped success feedback \\
Advantage estimation & Within-group RMS, no mean subtraction & Within-group RMS, no mean subtraction \\
KL coefficient & $0.02$ & $0.02$ \\
Advantage clip & $3$ & $3$ \\
Learning rate & $5\times10^{-6}$ & $5\times10^{-6}$ \\
Max sequence length & 16{,}384 & 16{,}384 \\
Evaluation interval & Every 25 updates & Every 25 updates \\
Training pool & 1{,}890 items & 2{,}432 games (four seen types) \\
Epochs / updates & 2 / 474 & --- \\
Sampled episodes & 30{,}240 & --- \\
\bottomrule
\end{tabular}}
\end{table}

\noindent\textbf{Post-training results.}
On ScienceWorld, GRPO optimization raises the average score to $66.9$ on strata175.
On ALFWorld, the main table reports each stage under its matched evaluation configuration. Knowledge-conditioned SFT obtains $90.3\%$, SFT followed by GRPO obtains $93.3\%$, and the separately marked External variant obtains $96.3\%$. GRPO uses only the four seen task types. Look and Pick2 remain unseen during parameter optimization; only the External variant permits test-time skill acquisition on these types.

\section{Extended Results and Statistical Analyses}
\label{app:results}

\noindent\textbf{Prompt-based evolution on ScienceWorld.}
Table~\ref{tab:sw_evolution} supplements the ablation results in Table~\ref{tab:saga_ablation} with the evolution schedule for the no-memory, trajectory-retrieval, and complete-SAGA configurations.
With DeepSeek-flash, the reported round-5 checkpoint from a six-round batch obtains $56.2$, compared with $25.0$ without memory and $54.8$ with direct trajectory retrieval.
With Qwen3.8-Max, the corresponding values are $48.4$, $63.7$, and $67.0$ after seven evolution rounds.
Both backbones exhibit the same rank ordering. Relative to trajectory retrieval, hierarchical memory yields absolute margins of $1.4$ points with DeepSeek-flash and $3.3$ points with Qwen3.8-Max.

\begin{table}[t]
\centering
\vspace{-1mm}
\caption{\small
Prompt-based evolution on ScienceWorld (strata175, strict configuration).
All conditions are evaluated on identical items.
The DeepSeek-flash row reports the round-5 checkpoint of a six-round evolution batch.
}
\label{tab:sw_evolution}
\scalebox{0.84}{
\begin{tabular}{llcc}
\toprule
Backbone & Condition & Avg. score & Evolution rounds \\
\midrule
\multirow{3}{*}{DeepSeek-flash}
& No memory & 25.0 & --- \\
& Trajectory retrieval ($L_0$) & 54.8 & --- \\
& SAGA ($L_0$--$L_3$) & 56.2 & 6 \\
\midrule
\multirow{3}{*}{Qwen3.8-max-0902}
& No memory & 48.4 & --- \\
& Trajectory retrieval ($L_0$) & 63.7 & --- \\
& SAGA ($L_0$--$L_3$) & 67.0 & 7 \\
\bottomrule
\end{tabular}}
\end{table}

\noindent\textbf{Utilization statistics.}
Table~\ref{tab:util_paired} reports supplementary episode-level paired analyses with Qwen2.5-7B-Instruct, examining the effects of direct knowledge injection and contextual instantiation.
Directly injecting the same $L_3$ knowledge yields a statistically significant negative effect, whereas contextual instantiation yields a statistically significant positive effect; the isolated instantiation contrast quantifies this reversal. Under instantiation, the estimated gain from principle-level knowledge is approximately three times that from procedural knowledge ($+0.0365$ versus $+0.0127$).
This analysis uses a harness distinct from the strict configuration in Table~\ref{tab:sci_config}. Because more than $80\%$ of episodes obtain zero terminal score under the 7B backbone, the paired estimates characterize the effect of knowledge utilization within this evaluation setting. They are reported as a separate supplementary analysis and are not directly comparable to the dual-backbone ablation scores in Table~\ref{tab:saga_ablation}.

\begin{table}[t]
\centering
\vspace{-1mm}
\caption{\small
Paired utilization analysis on ScienceWorld (Qwen2.5-7B-Instruct, 14 tasks, 3 seeds, $n{=}738$ episodes, scores on the 0--1 scale).
}
\label{tab:util_paired}
\scalebox{0.86}{
\begin{tabular}{lcc}
\toprule
Comparison & Paired difference & $t$ statistic \\
\midrule
Same $L_3$ knowledge, direct injection vs.\ no injection & $-0.0155$ & $-3.53$ \\
Same $L_3$ knowledge, instantiated vs.\ no injection & $+0.0365$ & $+5.90$ \\
Instantiation effect on $L_3$ (isolated) & $+0.0520$ & $+7.74$ \\
$L_2$ knowledge, instantiated vs.\ no injection & $+0.0127$ & --- \\
$L_3$ vs.\ $L_2$ under instantiation & $+0.0238$ & $+3.50$ \\
\bottomrule
\end{tabular}}
\end{table}

\noindent\textbf{Generalization to held-out ALFWorld task types.}
The ALFWorld protocol evaluates generalization to task types excluded from experience collection and parameter optimization.
The training pool contains Pick, Clean, Heat, and Cool, whereas Look and Pick2 are held out, comprising 18 and 17 evaluation games, respectively.
In every non-External configuration, neither held-out task type contributes trajectories to memory construction, supervised fine-tuning, or GRPO optimization; the memory is frozen before evaluation.
The Look and Pick2 columns in Table~\ref{tab:alfworld_main} therefore report performance on held-out task types within the same environment, while all evaluation scenes are drawn from \texttt{valid\_unseen}.
The External variant is reported separately because it permits test-time skill acquisition on these types and is not a zero-shot configuration.

\noindent\textbf{Round-wise evolution on ALFWorld.}
Table~\ref{tab:alf_rounds} reports the evolution process across rounds for the strong backbone and the two fixed-7B configurations.
With Qwen3.8-Max, SAGA improves the overall success rate from $96.3\%$ at R0 to $99.3\%$ at R3, corresponding to a net gain of four successful games among the 134 evaluation games. Relative to R1 ($97.8\%$), the gain at R3 corresponds to two additional successful games.
At R3, all 99 games from seen task types are successfully solved; the only remaining failure corresponds to an unseen Pick2 instance where the context budget is exhausted. No formatting failure is observed in any round.

For the fixed Qwen2.5-7B-Instruct actor, the memory-augmentation improvement from $45.5\%$ to $57.5\%$ discussed in the main text corresponds to the trajectory-only ($L_0$/$L_1$) configuration at R2. Its success rates are $45.5\%$, $46.3\%$, $57.5\%$, and $46.3\%$ at R0--R3, respectively.
The complete-hierarchy configuration is a separate experimental arm, with success rates of $45.5\%$, $48.5\%$, $54.5\%$, and $59.4\%$ at R0--R3, respectively. It improves across the reported rounds, although its R2 score remains below the trajectory-only control at the same round.
Thus, $57.5\%$ denotes the R2 trajectory-only result, whereas $59.4\%$ denotes the R3 result of the complete-hierarchy pipeline; these results should not be attributed to the same configuration. As stated in the table caption, no $L_3$ principle passed validation in the complete 7B configuration during R1--R3, so its improvement does not establish a contribution from validated principle-level knowledge.

\begin{table}[t]
\centering
\vspace{-1mm}
\caption{\small
Round-wise evolution on ALFWorld (\texttt{valid\_unseen}, 134 games, success rate in \%).
No principle from the complete 7B configuration passed validation in R1--R3.
}
\label{tab:alf_rounds}
\scalebox{0.86}{
\begin{tabular}{llcccc}
\toprule
Actor & Arm & R0 & R1 & R2 & R3 \\
\midrule
Qwen3.8-max-0902 & SAGA ($L_0$--$L_3$) & 96.3 & 97.8 & 97.8 & 99.3 \\
\midrule
\multirow{2}{*}{Qwen2.5-7B-Instruct}
& Trajectory only ($L_0$/$L_1$) & 45.5 & 46.3 & 57.5 & 46.3 \\
& Complete hierarchy & 45.5 & 48.5 & 54.5 & 59.4 \\
\bottomrule
\end{tabular}}
\end{table}

\noindent\textbf{GRPO trajectory on ScienceWorld.}
Starting from the SFT initialization, the dense-reward configuration raises the average score to $66.9$ on strata175. Figure~\ref{fig:grpo_evolution} shows the optimization trajectory over the course of training.
The supplementary material retains category-level scores and counts of perfect- and zero-score episodes.

\noindent\textbf{Mechanism of the improvement.}
Across the seven Qwen3.8-Max rounds, full hierarchical memory yields $11$--$14$ evaluation episodes with regression from a previously attained score, compared with $12$--$22$ under trajectory retrieval. The corresponding estimated rates of self-induced regression are $0.024$ and $0.045$, and the round-7 perfect-score counts are $41$ and $35$, respectively.
These diagnostics support the hypothesis that the principal observed benefit of knowledge utilization is reduced execution-time score regression, rather than an increased attainable score ceiling. This interpretation is consistent with the constraint-oriented mechanism described in Appendix~\ref{app:utilization}.

\section{Case Studies, Failure Analysis, and Limitations}
\label{app:cases}

\noindent\textbf{Knowledge formation and transfer in ScienceWorld.}
Figure~\ref{fig:hierarchy_example} presents the complete abstraction chain for a pigment-mixing task. Raw trajectories in which yellow and blue pigments produce green are summarized as episodic records, consolidated into a procedure for pigment selection, combination, and outcome verification, and lifted into a principle whose condition specifies the mixing of two primary pigments, whose effect is the generation of the corresponding secondary color, and whose mechanism is subtractive color mixing. The functional separation across levels is thereby made explicit: the principle abstracts a regularity that holds beyond any single color, whereas the procedure and episodes retain the operational specificity and evidential grounding of concrete executions.
This organization supports transfer without trajectory replay. For a task whose target color differs from the one observed during construction, such as orange rather than green, the principle is instantiated against the current goal and selects the red and yellow primaries for combination and subsequent verification, rather than reusing the green-producing trajectory retained at $L_0$ and $L_1$. The instantiated guidance constrains pigment selection at decision time, and a candidate action that selects a non-primary or off-target pigment is inconsistent with the principle and revised through the resampling step of Algorithm~\ref{alg:saga}; the episodic and procedural records remain available as fallback execution evidence.

\noindent\textbf{Directed behavior after evolution in ALFWorld.}
Under strong-backbone evolution, the three games that fail in round 1 (\texttt{clean/6}, \texttt{heat/29}, and \texttt{heat/103}) terminate after $48$, $38$, and $39$ steps, respectively, upon exhausting the context budget. At round 3, the same games are solved in $17$, $15$, and $13$ steps, respectively. These reductions are consistent with more directed exploration and cannot be attributed to an expanded interaction budget.
The only cross-round regression is \texttt{pick2/17}, which is solved within $37$ steps in round 1 but exhausts the context budget in subsequent rounds. This residual error occurs in the unseen Pick2 task type, for which no training trajectory is available. Table~\ref{tab:alf_case} summarizes these per-game episode lengths.

\begin{table}[t]
\centering
\vspace{-1mm}
\caption{\small Per-game episode length (steps at termination) before and after strong-backbone evolution in the strict ALFWorld setting. For the top three games, round 1 terminates at the context budget without success and round 3 solves the task; \texttt{pick2/17} is solved in round 1 but exhausts the budget thereafter.}
\label{tab:alf_case}
\scalebox{0.95}{
\begin{tabular}{lccc}
\toprule
Game & Round 1 & Round 3 & Outcome \\
\midrule
\texttt{clean/6}  & 48 & 17  & solved after evolution \\
\texttt{heat/29}  & 38 & 15  & solved after evolution \\
\texttt{heat/103} & 39 & 13  & solved after evolution \\
\texttt{pick2/17} & 37 & --- & regression (budget exhausted) \\
\bottomrule
\end{tabular}}
\end{table}

\noindent\textbf{Validation limitation in the complete-hierarchy arm with a fixed 7B actor.}
In the strict ALFWorld setting, no principle from the complete 7B configuration passes validation during R1--R3, as reported in Table~\ref{tab:alf_rounds}.
Despite this limitation, the complete-hierarchy pipeline improves from $45.5\%$ at R0 to $48.5\%$, $54.5\%$, and $59.4\%$ at R1, R2, and R3, respectively. At R2, the trajectory-only control obtains $57.5\%$, exceeding the complete-hierarchy result of $54.5\%$ by $3.0$ percentage points; at R3, the complete-hierarchy pipeline obtains $59.4\%$, compared with $46.3\%$ for the trajectory-only control.
These results distinguish task-solving performance from principle-validation outcomes: performance improves in the complete-hierarchy pipeline, but the reported rounds do not demonstrate successful validation or utilization of $L_3$ principles. Accordingly, we do not attribute the observed gains to validated principles or infer prompt-induced degradation from these aggregate scores alone.

\noindent\textbf{Limitations.}
The empirical conclusions are subject to several limitations.
First, both training-based pipelines use Qwen3.8-max-0902 for abstraction while optimizing Qwen2.5-7B-Instruct as the execution policy. This design does not isolate the contribution of the offline builder's pretrained knowledge; a same-model builder-actor control remains necessary.
Second, a substantial subset of residual ScienceWorld failures is execution-bound rather than knowledge-bound. Examples include episodes that locate the target object but fail to submit the final answer, which cannot be corrected through memory content alone.
Finally, the extracted principles are explanatory hypotheses supported by execution evidence, not established causal laws. Their generality beyond ScienceWorld and ALFWorld remains unverified.

\section{Future Work}
\label{app:future}

\noindent\textbf{Personalized experience adaptation.}
The present formulation treats memory as a task-centric artifact: episodes, procedures, and principles are abstracted from environment outcomes and shared across all interactions with a given benchmark. A natural extension is to condition the hierarchical memory on individual users, so that the same framework accumulates and reuses user-specific preferences, constraints, and interaction habits alongside task knowledge. Concretely, the source metadata of each node can be augmented with a user identifier, and retrieval can be biased toward evidence originating from, or validated against, the current user. This yields a personalized principle layer whose scope field is qualified not only by task type but also by user context, enabling the actor to instantiate constraints that respect stable individual preferences without re-deriving them at every episode.
Realizing this extension raises questions that the current evidence-grounding and utility-screening procedures do not yet address. User preferences are often implicit, occasionally contradictory, and may drift over time, so the validation criteria must distinguish durable preferences from transient instructions and must bound the influence of stale evidence. We regard controlled disentanglement of user-specific from task-general knowledge, together with per-user calibration of the injection budget, as the primary open problems for this direction.

\noindent\textbf{Interactive preference elicitation and lifelong personalization.}
A second direction couples memory construction with active elicitation. Rather than inferring preferences solely from passively observed trajectories, the actor could issue targeted clarification queries when retrieval returns conflicting or low-confidence user-conditioned principles, and record the resolved preference as a first-class episodic record. Such interactions would supply higher-quality supervision for the abstraction pipeline while keeping the human in the loop for genuinely ambiguous cases.
Sustaining personalization over long horizons additionally requires mechanisms for preference revision and forgetting. The exclusion registry introduced for rejected principles offers a starting point, but a personalized deployment must further support explicit user overrides, time-decayed evidence, and auditable provenance so that any behavioral change can be traced to a specific preference update. We leave a systematic study of these lifelong-personalization mechanisms, and their interaction with the training-based pipeline, to future work.

\end{document}